%% file: ms.tex
\documentclass{article}

\PassOptionsToPackage{numbers, compress}{natbib}

\usepackage[preprint]{neurips_2025}

\usepackage[utf8]{inputenc} 
\usepackage[T1]{fontenc}    
\usepackage{hyperref}       
\usepackage{url}            
\usepackage{booktabs}       
\usepackage{amsfonts}       
\usepackage{nicefrac}       
\usepackage{microtype}      
\usepackage{xcolor}         

\usepackage{amsmath}
\usepackage{graphicx}
\usepackage{tikz}
\usetikzlibrary{positioning}
\usetikzlibrary{shapes.geometric}
\usepackage{subcaption}
\newcommand\sect[1]{\S\ref{#1}}
\usepackage{algorithm}

\newcommand{\partitle}[1]{\vspace{1mm}\noindent\textbf{#1.}}

\title{Beyond Internal Data: Bounding and Estimating  Fairness from Incomplete Data}

\author{%
  Varsha Ramineni, Hossein A.~Rahmani, Emine Yilmaz, David Barber \\
  Centre for Artificial Intelligence\\
  University College London \\
  \texttt{\{varsha.ramineni.23, hossein.rahmani.22, emine.yilmaz, david.barber\}} \\
  \texttt{@ucl.ac.uk} \\
}

\begin{document}

\maketitle

\begin{abstract}
Ensuring fairness in AI systems is critical, especially in high-stakes domains such as lending, hiring, and healthcare. This urgency is reflected in emerging global regulations that mandate fairness assessments and independent bias audits. However, procuring the necessary complete data for fairness testing remains a significant challenge. In industry settings, legal and privacy concerns restrict the collection of demographic data required to assess group disparities, and auditors face practical and cultural challenges in gaining access to data. In practice, data relevant for fairness testing is often split across separate sources: internal datasets held by institutions with predictive attributes, and external public datasets such as census data containing protected attributes, each providing only partial, marginal information. Our work seeks to leverage such available separate data to estimate model fairness when complete data is inaccessible. We propose utilising the available separate data to estimate a set of feasible joint distributions and then compute the set plausible fairness metrics. Through simulation and real experiments, we demonstrate that we can derive meaningful bounds on fairness metrics and obtain reliable estimates of the true metric. Our results demonstrate that this approach can serve as a practical and effective solution for fairness testing in real-world settings where access to complete data is restricted.
\end{abstract}

\section{Introduction}
\label{sec:introduction}
It is well established that Artificial Intelligence (AI) systems have the potential to perpetuate, amplify, and systemise harmful biases \cite{buolamwini2018gender, caliskan2017semantics}. Therefore, rigorous testing for bias is imperative to mitigate harms, especially given the increasing influence of AI in high-stakes domains such as lending, hiring, and healthcare. Such concerns have fuelled active research on bias detection and mitigation \cite{mehrabi2021survey}, and ensuring the fairness of AI systems has become an urgent policy priority for governments around the world \cite{gov2023pro, aibill}. 
For instance, the EU AI Act imposes strict safety testing on high-risk systems \cite{aiact}, while New York City Local Law 144 mandates independent bias audits for AI used in employment decisions \cite{Groves2024-xo}. 

Despite this progress, procuring the necessary data for fairness testing remains a significant challenge. Influential works in ethics and fairness of machine learning have highlighted the centrality of datasets \cite{jo2020lessons,bao2021s}, emphasising how representative  model testing and evaluation data is crucial \cite{bergman2023representation,shome2024data}. 
To effectively uncover biases, complete datasets that include demographic information and their relationship with model features are essential. However, having access to such datasets that can reliably be used for evaluating fairness may not always be possible in practice. 

As a motivating example, consider a bank that uses an AI system to assess loan applicants based on non-protected variables such as \textit{savings} and \textit{occupation}. The bank wants to perform an \emph{internal} audit as to whether its AI system inadvertently discriminates against certain ethnic groups. Most common fairness metrics assess disparities in decision outcomes across protected attribute groups, such as \textit{ethnicity}. Therefore, the bank requires protected attribute data of the applicants, alongside data of non-protected attributes required by the model to make a loan decision. 

\begin{figure}[t]
    \centering
    \includegraphics[width=0.9\linewidth]{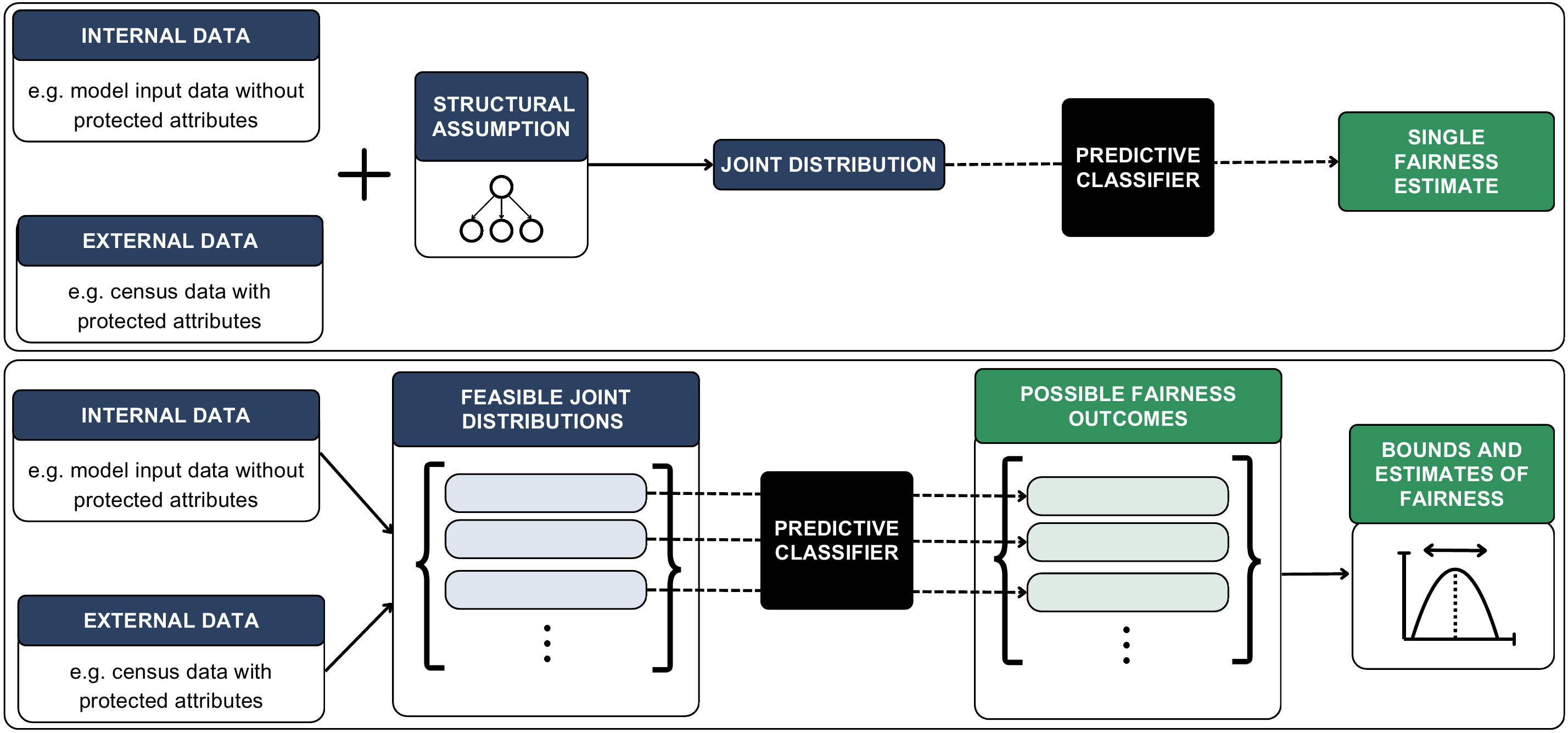}
    \caption{Illustration of our approach to bound and estimate fairness from incomplete data. The internal dataset (e.g., from a bank) contains non-protected attributes like savings and occupation, but lacks protected demographic attributes. A separate external public dataset includes protected attributes (such as ethnicity) and overlaps partially with the internal dataset (e.g., occupation attribute is common in both datasets). (\textbf{Top}) The first approach assumes a strong structural independence assumption to produce a single estimated joint distribution, and therefore a single fairness estimate. (\textbf{Bottom}) The second approach uses the marginal distributions from both sources to estimate the set of joint distributions consistent with the observed marginals. This allows fairness metrics to be computed over the space of feasible distributions, enabling bounding and estimation of fairness even in the absence of complete data.}
    \label{fig:approach-diagram}
\end{figure}

Although protected attributes such as ethnicity, sex, age etc.~are crucial to assess bias, their collection and use in modelling are heavily restricted under regulations such as GDPR \cite{andrus2021we,veale2017fairer}. Hence, most internal datasets collected by organisations that use AI systems for decision making (such as the bank in our example) do not contain such protected attributes \cite{madaio2022assessing}. Similarly, procuring the necessary data is also a huge complexity for auditors, hindering the effective implementation of algorithmic auditing laws \cite{Groves2024-xo}. In an \emph{external} audit of fairness, the auditing agency often has access only to the black box loan predictions and is not provided any data by the bank since existing regulations often do not allow data holders to release datasets that pause privacy concerns. For this external audit the agency needs a joint distribution of both the attributes needed by the black box loan classifier and the protected attributes. Therefore, the development of principled testing approaches capable of effectively uncovering biases in limited data scenarios is essential \cite{madaio2022assessing}.

Our work addresses this challenge of estimating classifier fairness in scenarios where complete data including protected attributes is inaccessible. In practice, data relevant for fairness testing is often split across separate sources, such as internal datasets held by institutions and external public datasets each providing only partial, marginal information. We propose leveraging such separate data with overlapping variables, which are more accessible than complete datasets containing all variables \cite{frogner2019fast}. Specifically, in addition to using an internal dataset that lacks protected attribute information, we propose utilising external public data, such as census datasets which provide representative demographic information. For example, the UK Office for National Statistics \cite{ONS2021Census} offers multivariate data from the 2021 Census, providing access to customisable combinations of census variables. Similar resources are available from the US Census Bureau \cite{USCensus}. In our motivating example above,  the internal dataset used by the bank only includes information about non-protected attributes $\{ \textit{savings}, \; \textit{occupation}\}$. Therefore to estimate fairness it is valuable to utilise an external public dataset such as  $\{ \textit{occupation}, \; \textit{ethnicity}\}$ that contains required protected attribute information and is representative of the population where the model will be deployed.

As illustrated in Figure~\ref{fig:approach-diagram}, we propose an approach to leverage available separate incomplete data to estimate the set of all joint distributions consistent with the available marginal information. This enables us to compute fairness metrics over the full range of feasible distributions, allowing for estimation and bounding of fairness even when protected attributes are missing from the model’s input data. This is important to robustly estimate fairness metrics, as the true distribution underlying the data is often unknown or difficult to determine. We validate our approach through simulation and real experiments with realistic scenarios of separated data with limited overlap. Our results demonstrate that we can derive meaningful constraints on fairness metrics and obtain reliable estimates of the true metric, highlighting the effectiveness of our method in evaluating fairness when complete data is unavailable. These findings suggest that our approach offers a robust and practical solution for fairness evaluation in contexts where full datasets are inaccessible. It provides organizations with a viable means to assess fairness, helping them meet ethical and legal requirements even when data is limited.

\section{Methodology}

Returning to our motivating example, the bank that uses a predictive model $p(\hat{y} \mid \textit{savings}, \textit{occupation})$ to determine loan approvals, where $\hat{y} \in \{0, 1\}$ indicates whether a loan is granted depending on the non protected attributes such as applicant’s salary and occupation. We would like to assess whether this prediction is fair against a protected attribute \textit{ethnicity}. Many standard fairness metrics, such as Demographic Disparity and Disparate Impact \cite{mehrabi2021survey}, are defined in terms of differences in the model's output across groups defined by protected attributes. Therefore, calculating the conditional distribution $p(\hat{y} \mid \textit{ethnicity})$ is central to evaluating fairness. We can express this conditional distribution as:

\begin{align}
p(\hat{y} \mid \textit{ethnicity}) = \sum_{\textit{savings, occupation}} p(\hat{y} \mid \textit{savings, occupation}) \, p(\textit{savings, occupation} \mid \textit{ethnicity}).
\label{eq:py_given_e}  
\end{align}

Here, $p(\hat{y} \mid \textit{savings, occupation})$ is fixed by the bank’s loan prediction model, and the key challenge is to estimate the required distribution $p(\textit{savings, occupation} \mid \textit{ethnicity})$,  which depends on the joint distribution 

\begin{equation}
p(\textit{savings, occupation, ethnicity}).
\end{equation} 

However, this distribution is not directly available, as the bank only observes the marginal $p_{\text{bank}}(\textit{savings, occupation})$, and public datasets such as census data include protected attributes but often as marginals such as $p_{\text{public}}(\textit{occupation, ethnicity})$, where there could be at least one overlapping variable with the internal data. 

This leads to the central question: \textit{How can we bound and estimate fairness metrics using only the available marginals such as $p_{\text{bank}}(\textit{savings, occupation})$ and $p_{\text{public}}(\textit{occupation, ethnicity})$?} Access to the joint distribution of protected attributes and predictive features is a central requirement for fairness testing, yet it is often unavailable in practice. Therefore, estimating it accurately from available data is essential for reliable fairness evaluation.



\subsection{Problem Setup and Notation}

In the general case, our goal is to enable fairness evaluation by estimating the required joint distribution,
\begin{equation}
p(\textit{internal}, \textit{common}, \textit{external}),
\end{equation} 

using only two incomplete marginal datasets with overlap between them (internal and external data detailed below). The \emph{common} variable denotes the one shared by both datasets, serving as the overlapping link (e.g., occupation). Note that \textit{internal}, \textit{common}, and \textit{external} variables could also each represent a set of multiple variables, all derivations generalise accordingly. 

\begin{itemize}
    \item \textbf{Internal Data} \( p_{\text{i}}(\textit{internal}, \textit{common}) \), from a private or institutional source (e.g., bank data), typically containing predictive features (e.g., savings, occupation).
    \item \textbf{External Data} \( p_{e}(\textit{common}, \textit{external}) \), from a public or population-level dataset (e.g., census data), typically containing socio-demographic information and protected attributes  (e.g., occupation, ethnicity)
\end{itemize}

In this work, we explore two contrasting approaches for estimating the full joint distribution from the incomplete marginal data. The first relies on a strong structural assumption to produce a single estimated joint distribution. The second makes no additional assumptions and instead explores the entire set of feasible joint distributions that align with the marginals. These two approaches are illustrated in Figure~\ref{fig:approach-diagram} and are detailed below.

\subsection{Structural Assumption Method}
\label{sec:structural-assum}
Reconstructing a full joint distribution from partial marginals is inherently ill-posed, as multiple joint distributions may be consistent with the observed data. Therefore it requires additional assumptions. One potential approach to estimate the joint distribution is to use marginal data observations along with a structural independence assumption, applying techniques like maximum likelihood estimation. Once the joint distribution is estimated, it can be used to compute a single fairness metric estimate using Equation~\ref{eq:py_given_e}, as illustrated in Figure~\ref{fig:approach-diagram}. We consider below two simple structural independence assumptions to fit a joint distribution.

\partitle{Latent Na\"ive Bayes} We employ a latent variable model based on the Naïve Bayes assumption by introducing a latent variable $z$, which assumes that all variables in the joint are conditionally independent given $z$. We use the Expectation-Maximization (EM) algorithm \cite{dempster1977maximum} to train the model (see Appendix~\ref{app:em-algo} for details).

\begin{equation}
p(\textit{internal, common, external}) = \sum_z p(z) \prod_{x \in \{ \textit{internal} \cup \textit{common} \cup \textit{external} \} }  p(x \mid z)
\label{eq:latent_nb}
\end{equation}

\partitle{Marginal Preservation}
We preserve one of the observed marginals, while approximating the other through conditional probabilities. For example, we could preserve the external public data.

\begin{equation}
p(\textit{internal, common, external}) = p_e(\textit{common, external}) \, p_i(\textit{external} \mid \textit{common})
\end{equation}

\subsection{Feasible Set Method}
Rather than relying on an assumption to obtain a single feasible joint distribution, a more robust approach could be to estimate the entire feasible set of joint distributions that satisfy the constraints imposed by the available marginal distributions. Below, we detail two cases for defining feasible joint distributions based on the available marginal datasets, depending on whether the marginals are consistent, i.e., come from same underlying distribution, or in practical terms depending on whether the datasets represent similar populations.

Since a given predictive classifier is to ideally be evaluated on data from a representative population, we assume that the external data distribution (e.g. census data) should be preserved i.e. $p_{\text{e}}(\textit{common}, \textit{external}) = \sum_{\textit{internal}} p(\textit{internal}, \textit{common}, \textit{external})$.

\subsubsection{Constraints for Consistent and Inconsistent Marginals}  
The first case we consider is \emph{marginal consistency}, where the available marginal distributions (e.g., \(p_{\text{i}}(\textit{internal}, \textit{common})\) and \(p_{\text{e}}(\textit{common}, \textit{external})\)) come from the same underlying joint distribution. Specifically, we require a joint distribution \(p(\textit{internal}, \textit{common}, \textit{external})\) satisfying:
\begin{align}
p_{\text{e}}(\textit{common}, \textit{external}) &= \sum_{\textit{internal}} p(\textit{internal}, \textit{common}, \textit{external}), \\ p_{\text{i}}(\textit{internal}, \textit{common}) &= \sum_{\textit{external}} p(\textit{internal}, \textit{common}, \textit{external}).
\end{align}
From these, it follows that the marginal distribution over the overlapping \emph{common} variables must match,  $p_{i}(\textit{common}) = p_{e}(\textit{common})$
This assumption is reasonable when both datasets represent the same underlying population.

In practice, exact marginal matching \(p_{\text{i}}(\textit{common}) = p_{\text{e}}(\textit{common})\) may not hold. In this case, we relax the marginal consistency requirement and instead seek a joint distribution that closely aligns with the available marginals under reasonable constraints. In this case, we may still require full consistency with the external data, but only enforce that the model’s conditional distribution matches the internal data conditionals:
\begin{align}
p_{\text{i}}(\textit{internal} \mid \textit{common}) &= \frac{\sum_{\textit{external}} p(\textit{internal}, \textit{common}, \textit{external})}{\sum_{\textit{internal}, \textit{external}} p(\textit{internal}, \textit{common}, \textit{external})}.
\end{align}




\subsubsection{Bounding and Estimating Fairness using Feasible Set}
\label{method:bounds}


In both cases, the problem ultimately reduces to obtaining a feasible joint distribution that satisfies the given marginal constraints. 
As shown in Figure~\ref{fig:approach-diagram}, given two separate marginal datasets, we could obtain a list of feasible set of joint distributions $\mathcal{P}$, where each $p^i \in \mathcal{P}$ satisfies the marginal constraints.

\begin{equation}
\mathcal{P} = \{ p^1(\textit{internal, common, external}), p^2(\textit{internal, common, external}), \dots \}
\end{equation} 

Then for each $p^i \in \mathcal{P}$, along with the fixed predictive classifier probabilities $p(\hat{y} \mid internal, common)$, we derive the corresponding conditional probability $p(\hat{y} \mid e)$ defined in Equation~\ref{eq:py_given_e}, for a chosen protected attribute $e$. This enables us to compute fairness metrics such as Disparate Impact (defined in Equation~\ref{eq:di}), which is a function of $p(\hat{y} \mid e)$. Therefore we are able to obtain a list of possible fairness metrics $
\mathcal{M} = \{ m^1, m^2, ...\} $,
computed across the set of feasible joint distributions. By identifying the maximum, minimum, and average values of the set 
$ \mathcal{M} $, we can derive bounds and estimates on the fairness metric. These bounds would capture the full range of possible fairness outcomes, given the available data and  constraints. 

To first obtain the feasible set of joint distributions $\mathcal{P}$ given two marginal distributions, there are many possible methods \cite{frogner2019fast}. In this paper, we adopt a simple approach under the assumption that all variables are binary. We represent the full joint distribution as a finite-dimensional vector, where each entry corresponds to the probability of one combination of variable values. This vector must satisfy the follow constraints: all entries must be non-negative, they must sum to one (ensuring a valid probability distribution), and they must satisfy the observed marginal constraints. Since the system is underdetermined, we systematically vary the free parameters within their valid ranges to generate all feasible joint distributions. This procedure is outlined in detail in Appendix~\ref{app:joint}. While this method is theoretically extendable to a larger number of variables, the computational complexity grows exponentially with the number of variables (see Appendix~\ref{app:general-case}). Our experiments focus on estimating joint distributions consisting of three binary variables which: two predictive attributes and one protected attribute. This study therefore serves as an exploration of our proposed approach, demonstrating its feasibility and highlighting directions for further research.

\section{Simulated Study: Feasible Set Method Across Wide Range of Scenarios}
We conduct simulation experiments to evaluate how well fairness metrics can be bounded and estimated using the feasible set method (approach shown in Figure~\ref{fig:approach-diagram}) under a wide range of conditions. All our code and results are available at \url{https://github.com/varsharamineni/fairness-incomplete-data}.

\partitle{Ground Truth And Predictive Classifier} We start by sampling a ground truth joint distributions over three binary variables from a Dirichlet prior. The ground truth serves as the reference from which true fairness metrics are computed.  

\partitle{Marginal Data with Varying Inconsistency} We wish to simulate the limited data scenario where complete data is not available. We experiment with both consistent marginals (perfectly derived from the ground truth) and inconsistent marginals by introducing controlled distortion via scaling factors $\alpha$ and $\beta$, mimicking real-world data mismatches. To quantify the degree of mismatch between the marginal distributions, we use the Kullback–Leibler (KL) divergence between the common variable's marginals. 

\partitle{Predictive Classifier} To simulate a fairness testing scenario, we generate classifier predictions by sampling conditional probabilities of outcomes given predictive features (from non-overlapping variables in external marginal data) from a Dirichlet distribution, representing a generic model under evaluation.

We repeat this process to generate 1,000 distinct ground truth distributions and predictive classifiers. These are paired with 25 marginal datasets (1 consistent and 24 inconsistent), resulting in a total of 25,000 unique fairness testing scenarios under limited data.

\partitle{Comparing Feasible Set Method with Ground Truth} For each pair of marginal distributions, we obtain a list of feasible joint distributions that meet the marginal constraints. On average, this set contains approximately 5,951 joint distributions, with a maximum of 10,000. We then compute the fairness metrics for each feasible joint to obtain a range of possible fairness values. We then compare these possible values to the true metric obtained from the ground truth distribution.

\begin{figure}[t]
    \centering
    \includegraphics[width=0.9\textwidth]{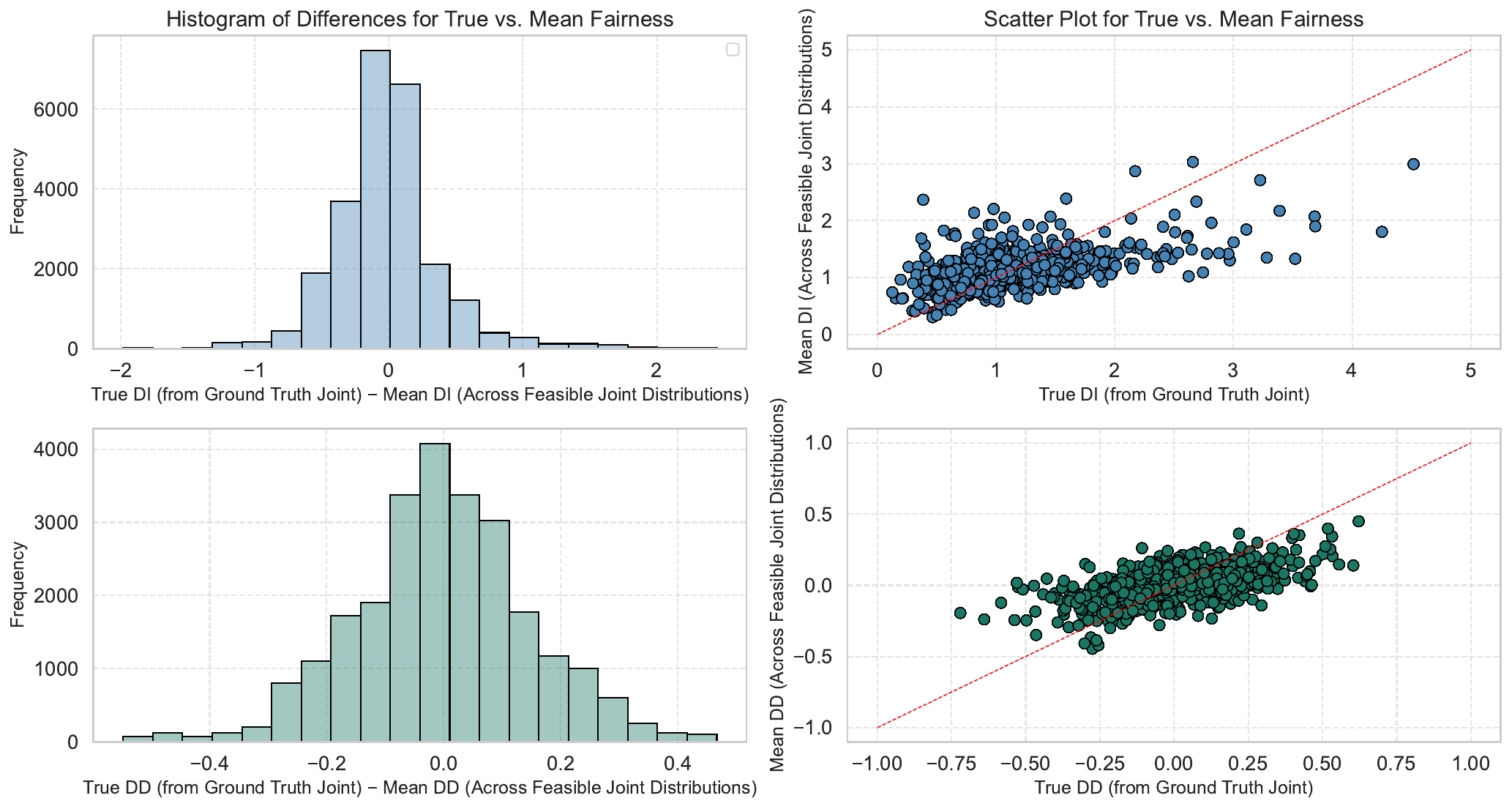}
    \caption{
       Comparison of mean fairness metrics derived from feasible joint distributions with true fairness metrics computed from the ground truth joint across 25,000 simulated fairness testing scenarios using limited marginal data. 
       Left: Histogram showing the distribution of differences between the true fairness metric and the mean fairness metric across feasible joint distributions.
       Right: Scatter plot comparing true fairness metrics against the mean estimates, with the top row illustrating Disparate Impact (DI) and the bottom row showing Demographic Disparity (DD). The red dashed line represents perfect agreement between the two.
    }
    \label{fig:fairness-comparison}
\end{figure}

\subsection{Fairness Metrics}

We focus on the Disparate Impact (DI) metric, which measures the relative rate of favourable outcomes (e.g., loan approval) between unprivileged and privileged groups. Letting $e$ denote the protected attribute (e.g., ethnicity), DI is computed as:

\begin{equation}
\text{DI} = \frac{p(\hat{y} = 1 \mid e = 0)}{p(\hat{y} = 1 \mid e = 1)}
\label{eq:di}
\end{equation}

The DI metric theoretically ranges from $0$ to $\infty$: values below $1$ indicate disadvantage to the unprivileged group, values above $1$ indicate advantage, and $1$ corresponds to perfect parity between groups. In practice, extremely large values of DI may indicate infeasible distributions therefore we remove scenarios leading to DI greater than 5. 

We further focus on Demographic Disparity (DD) which measures the difference in favourable outcomes between unprivileged and privileged groups.

\begin{equation}
\text{DD} = p(\hat{y} = 1 \mid e = 0) - p(\hat{y} = 1 \mid e = 1)
\end{equation}

While both metrics rely on estimating $p(\hat{y} \mid e)$, they respond differently to probability shifts. For instance, since DI is a ratio, even small shifts in the conditional distribution can lead to large changes in the ratio.

\subsubsection{Results: Average Fairness Metric from Feasible Set Method}


Our evaluation covers 25,000 different marginal dataset pairs, where for each pair we compare the true fairness metric (calculated from the known ground truth joint distribution) to the average fairness metric across all feasible joint distributions.

Figure~\ref{fig:fairness-comparison} shows the distribution of differences between the true and average DI and DD values across the different 25,000 marginal pairs, with varying inconsistency levels. Both distributions are centred around zero, with mean differences of $-0.010, (\pm 0.410)$ for DI and $-0.002, (\pm 0.151)$ for DD. This indicates that the average of plausible metrics across the feasible joints provides a slight underestimation of the true fairness values on average. However, this bias is small in magnitude, suggesting that the average over possible fairness estimates provides a reasonably accurate approximation of the ground-truth fairness metric in most cases.

\subsubsection{Results: Range of Fairness Metrics and Bounding of True Metric}

In our experiments, we observe that $100\%$ of the derived bounds from the possible fairness values across feasible joint distributions consistently contain the true fairness metric. This demonstrates that even with our simple estimation approach to generate feasible joint distributions (described Section~\ref{method:bounds} and detailed in Appendix~\ref{app:joint}), we can achieve accurate bounds. We further evaluate our feasible set based approach by examining the tightness of the fairness bounds produced from the set of feasible joint distributions consistent with the incomplete marginal data. Across 25,000 scenarios, the average range is $1.828, (\pm 3.211)$ for DI and $0.569, (\pm 0.302)$ for DD. While these bounds reliably contain the true metric, many of the feasible joint distributions may not be realistic in practice, highlighting the potential benefit of incorporating additional external information to further tighten the bounds.

\subsection{Results: Impact of Marginal Inconsistency}

Across our 25,000 scenarios, we varied inconsistency between the marginal datasets to assess our approach, with KL divergence values ranging from 0 (perfectly consistent) to 2 (highly inconsistent).
We investigated how fairness estimation performance varies across different levels of marginal inconsistency. We observe that across all levels of marginal inconsistency (0–0.01, 0.01–0.05, 0.05–0.1, 0.1–0.3, 0.3–0.5, 0.5+), of marginal inconsistency, the average difference between the true fairness metric and the mean over feasible joint distributions remains small, below $0.003$ for DD and around $0.01$ for DI. This suggests that the average over the possible set provides a reasonably accurate estimate of the true fairness value, even when the marginal data is inconsistent. The range of possible fairness values remains stable across bins: approximately $0.56 - 0.57$ for DD and $1.67 - 1.91$ for DI. This suggests that while uncertainty is present due to incomplete data, the bounds do not deteriorate substantially with increasing inconsistency. Together, these results show that our method offers reliable and informative fairness estimates, even under significant marginal mismatch. 

\section{Real-World Data Experiments}

\begin{table}
  \caption{Overview of real-world datasets used in experiments}
  \label{tab:data}
  \centering
  \begin{tabular}{lllll}
    \toprule
    \textbf{Name}     & \textbf{\# Instances}     & \textbf{\# Attributes}  & \textbf{Label} & \textbf{Protected Attribute} \\
    \midrule
    Adult \cite{adult_2} & 45,222  & 13  & Income & Sex (67.5\% male, 32.5\% female)  \\
    COMPAS \cite{compas} & 5278  & 9  &  Recidivism  & Race (60.2\% white, 39.8\% black) \\
    German \cite{dheeru2017uci}    &  1000       & 22 & Credit Risk & Sex (69\% male, 31\% female)\\
    \bottomrule
  \end{tabular}
\end{table}

\begin{figure}[t]
\centering
    \includegraphics[width=0.9\textwidth]{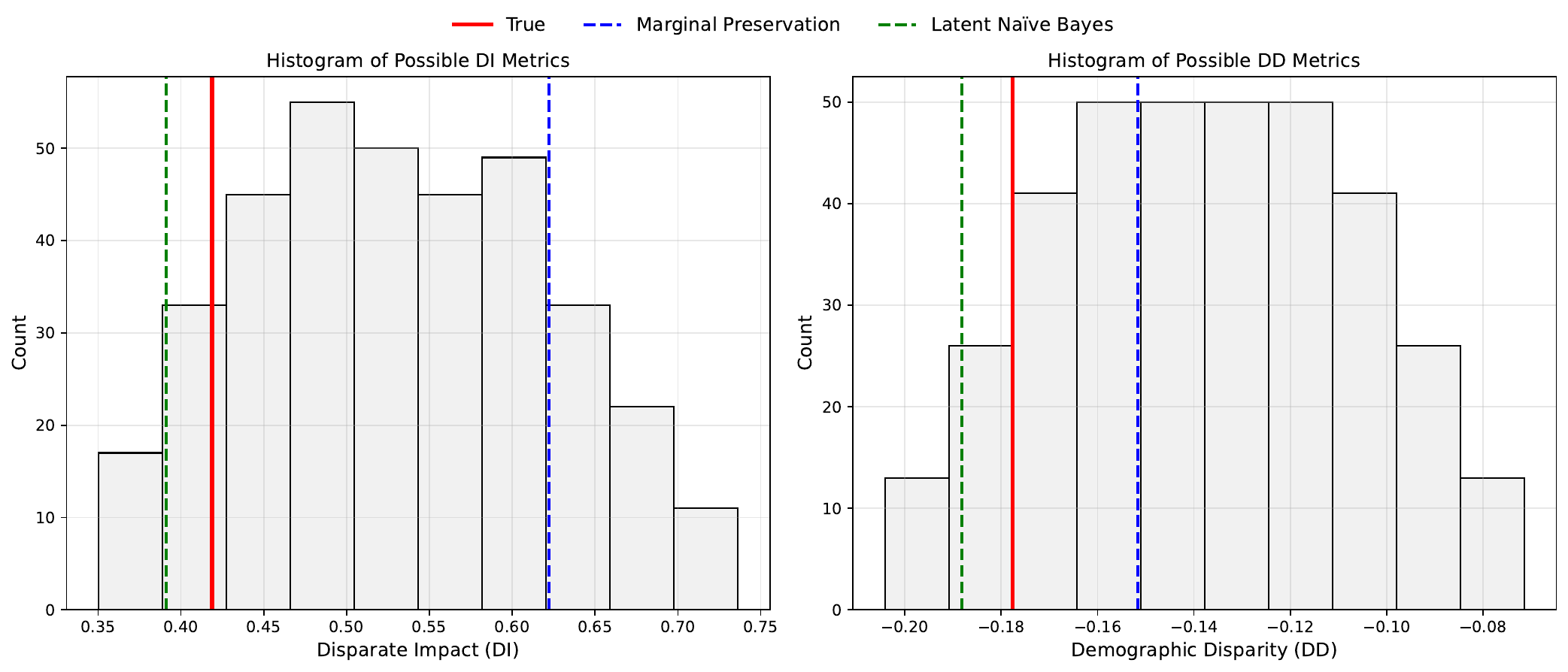}
    \caption{Distribution of possible values for a fairness metric, Disparate Impact (DI) or Demographic Disparity (DI), computed from the feasible set of joint distributions consistent with observed marginals. The true fairness metric (from the full joint) is marked in red. Fairness estimates from using two structural assumption-based methods are shown in green and blue for comparison. Results using Adult Data. }
    \label{fig:real_hists}
\end{figure}

We conduct our experiments using three real-world datasets: \textbf{Adult} \cite{adult_2}, \textbf{COMPAS} \cite{compas}, and \textbf{German Credit} \cite{dheeru2017uci}, detailed in Table~\ref{tab:data}, which are commonly used in the fairness literature \cite{le2022survey}. For all three datasets we follow the literature by removing instances with null values, and map all continuous variables into categorical variables (see Appendix~\ref{app:real_data_details} for details) \cite{le2022survey}. These datasets represent complete real data with protected attributes. Our goal is to approximate this joint data distribution from incomplete data.  In our experiments we simulate the scenario where only two separate marginal datasets are available. We design these carefully to create realistic scenarios of internal and external datasets (making sure the overlapping variable chosen is likely to be in census data, e.g. housing). We also wish to simulate having a trained classifier that we wish to test for fairness. This is done by training a decision tree classifier models on variables in the internal dataset, which does not include protected attributes (this data is not used to estimate the joint). Full details along with data splits are in Appendix~\ref{app:real_data_exp}.

\partitle{Evaluating Structural Assumption Method}
We use the full datasets with all variables for these experiments and also keep all the multivariate categories. Due to the number of parameters present in the joint, we using sampling from the datasets directly to estimate the joint distribution using the two structural independence assumptions: Latent Naive Bayes and Marginal Preservation. Therefore we get a joint dataset, from which we bootstrap to account for sampling variability, and and use to test the classifier and obtain classifier fairness.

\partitle{Evaluating Feasible Set Method}
We reduce each dataset to three binary variables to match the feasible set framework: two marginal distributions (e.g., internal and external datasets) with one overlapping variable. Using these marginals, we construct the feasible set of joint distributions the set of all joint distributions consistent with the observed marginals. From this feasible set, we compute a range of possible values for fairness. We also compute: The true fairness metric using the full joint data distribution (ground truth), as well as fairness estimates from two single joint estimation methods based on structural assumptions (detailed in Section~\ref{sec:structural-assum}) shown for comparison.

\begin{table}[t]
  \caption{Absolute difference in Disparate Impact (DI) between fairness metrics derived from using joint estimated using two structural independence assumptions and the true joint (Latent Bayes, and Marginal Preservation) for a Decision Tree Classifier.}
  \label{tab:di_only}
  \centering
  \begin{tabular}{llc}
    \toprule
    \textbf{Dataset – Attribute} & \textbf{Method } & \textbf{DI diff} $\downarrow$ \\
    \midrule
    Adult – Sex & Latent  & 0.162 \\
                & Marginal & 0.047 \\
    \midrule
    COMPAS – Race & Latent  & 0.001 \\
                  & Marginal & 0.000 \\
    \bottomrule
  \end{tabular}
\end{table}

\subsection{Results and Practical Implications}

We present key findings in Table~\ref{tab:di_only} and Figure~\ref{fig:real_hists}, comparing fairness estimates derived from structural assumptions to those obtained using the feasible set approach. In Figure~\ref{fig:real_hists}, the true fairness value, along with estimates from Latent Naïve Bayes and Marginal Preservation (blue and green), consistently fall within the bounds of the feasible set. This is expected, as these structural assumptions are consistent with the observed marginals. Table~\ref{tab:di_only} highlights that the structural assumptions yield accurate estimates in the COMPAS dataset, where the assumptions may plausibly hold, but are less reliable in the Adult dataset, highlighting the limitation.

The DI metric is frequently used in legal and regulatory frameworks to assess fairness, particularly in employment, credit, and housing decisions. According to the U.S.~Equal Employment Opportunity Commission (EEOC) \cite{eeoc2007tests}, a DI value below 0.8 may indicate discriminatory impact. With the derived bounds shown in Figure~\ref{fig:real_hists}, we see that all possible values from the data are below 0.8, which could be valuable to show regulators or as an early internal warning sign.

\section{Related Work}
\partitle{Fairness Testing} Significant work on fairness evaluation has centered on formalising definitions of fairness \cite{mehrabi2021survey} and emphasising the critical role of data \cite{ bao2021s, jo2020lessons, gebru2021datasheets, paullada2021data}. Recent work has also explored fairness testing in response to regulatory requirements \cite{Groves2024-xo, veale2017fairer} and in the context of industry \cite{Holstein2019-gt, madaio2022assessing} and software development \cite{chen2021synthetic}. Additionally, there is growing interest in sample-efficient approaches to fairness testing \cite{JiSS20, van2024can}.

\partitle{Fairness without Protected Attributes}
Approaches to fairness without demographics include proxy methods and attribute classifiers to infer sensitive attributes, though these often introduce bias and legal concerns \cite{zhu2023weak, ashurst2023fairness}. Some propose disclosure for trusted third parties to collect sensitive attributes \cite{veale2017fairer}, but access to such data remains challenging in practice \cite{Groves2024-xo}. Several closely related studies focus on a similar data setting as ours, using auxiliary public data \cite{kallus2022assessing, elzayn2024estimating}, highlighting that this separate data access scenario is common and realistic. Their theoretical contributions greatly complement our work. However, our focus is on accurately estimating the joint distribution. Unlike prior works, we provide a simple method for explicitly obtaining the full set of feasible joint distributions, and demonstrate that meaningful fairness estimation is still possible even when the observed marginals are inconsistent. We further offer a comparison between two fundamentally different estimation strategies.

\section{Conclusion and Future Work}
In this study, we addressed the challenge of evaluating classifier fairness in settings where complete datasets, including protected attributes, are unavailable due to practical and privacy concerns. We proposed a method that leverages separate, overlapping datasets, commonly encountered in real-world scenarios, to estimate the joint distribution required for fairness evaluation. We investigated two contrasting approaches for estimating the required joint distribution from incomplete marginal data: one based on a strong structural assumption (e.g. Latent Naive Bayes) to derive a single joint estimate, and another assumption-free method that explores the full set of feasible joint distributions consistent with the observed marginals. 

Our findings demonstrated that averaging over the feasible set provides fairness estimates that closely approximate the true metric, indicating low estimation bias. Even with the simple estimation method for obtaining the feasible set, our bounds consistently contained the true values. Moreover, low bias and the range of possible fairness values remains relatively stable across different levels of marginal inconsistency, highlighting the robustness of our bounds-based approach.

This work has promising practical applications, for both industry and regulators, offering
a viable substitute for fairness testing when complete data is inaccessible. While our simulations for the feasible set method were limited to three variables, future work should generalise to higher-dimensional settings and develop efficient strategies for sampling the feasible space. The observed stability in the range of possible fairness values may be influenced by the limited number of variables. Additionally, we focused exclusively on one type fairness measurement; evaluating the applicability of our approach to other fairness definitions remains an important direction for future research.


\begin{ack}
The authors declare no competing interests related to this paper. This research was supported by the UKRI Engineering and Physical Sciences Research Council (EPSRC) [grant numbers EP/S021566/1 and EP/P024289/1].
\end{ack}

{
\small 
\bibliography{references}
}


\medskip


\newpage

\appendix

\section*{Technical Appendices and Supplementary Material}

\input{appendix}


\newpage

\end{document}

%% file: appendix.tex
\numberwithin{figure}{section}
\numberwithin{table}{section}
\numberwithin{equation}{section}

\paragraph{Appendix} This appendix accompanies the paper \textit{Beyond Internal Data: Bounding and Estimating Fairness from Incomplete Data}.

It is organised as follows:

\paragraph{\sect{app:joint}} provides technical details on the feasible set method, which focuses on obtaining the set of joint distributions consistent with observed marginals, as introduced in the main text

\paragraph{\sect{app:general-case}} extends the feasible set method to the general case, with additional discussion on computational complexity

\paragraph{\sect{app:em-algo}} presents technical details of the Latent Naïve Bayes model, a structural assumption based method used for joint distribution estimation, as introduced in the main text

\paragraph{\sect{app:real_data_details}} describes the real-world datasets used in the experiments, including tables detailing the variables

\paragraph{\sect{app:real_data_exp}} describes the experimental setup and details for the real-world datasets, including additional results omitted from the main text

All code and data are publicly available at: \url{https://github.com/varsharamineni/fairness-incomplete-data}

\section{Obtaining Feasible Set of Joint Distributions}
\label{app:joint}

\subsection{Case For Inconsistent Marginals}
\label{app:incost_marg}

We are given two marginal distributions: 
\(p_\text{bank}(s, o)\) and \(p_\text{public}(o, e)\), where: \(s, o, e \in \{0, 1\}\) are binary variables. Our goal is to obtain the set of feasible joint distributions \(p(s, o, e)\) that align with the following constraints detailed below.

\subsubsection*{Constraints}

Let \(p_{soe} = p(s, o, e)\). Then the distribution is subject to the following constraints:

\begin{itemize}
    \item \textbf{Non-negativity:}
    \begin{align}
        p_{soe} \geq 0, \quad \forall s, o, e \in \{0, 1\}
    \end{align}

    \item \textbf{Normalisation:}
    \begin{align}
        \sum_{s, o, e} p_{soe} = 1
    \end{align}

    \item \textbf{Marginal Constraints from \(p_\text{bank}(s| o)\)}
    \begin{align}
    (p_{000} +p_{001}+p_{100}+p_{101} )p_\text{bank}(0\mid0)&=p_{000}+p_{001},\\
    (p_{000} +p_{001}+p_{100}+p_{101} ) p_\text{bank}(1\mid0)&=p_{100}+p_{101},\\
    (p_{010} +p_{011}+p_{110}+p_{111})p_\text{bank}(0\mid1)&=p_{010}+p_{011},\\
    (p_{010} +p_{011}+p_{110}+p_{111})p_\text{bank}(1\mid1)&=p_{110}+p_{111}.
    \end{align}
  
    \item \textbf{Marginal Constraints from \(p_\text{public}(o, e)\):}
    \begin{align}
        p_{000} + p_{100} &= p_\text{public}(0, 0) \\
        p_{010} + p_{110} &= p_\text{public}(1, 0) \\
        p_{001} + p_{101} &= p_\text{public}(0, 1) \\
        p_{011} + p_{111} &= p_\text{public}(1, 1)
    \end{align}
\end{itemize}

Note that the two marginals themselves sum to 1:
\begin{equation}
\sum_{s, o} p_\text{bank}(s, o) = 1, \quad \sum_{o, e} p_\text{public}(o, e) = 1
\end{equation}

\subsubsection*{Removing Redundancy}

There are six non-redundant equations in total, accounting for the fact that the marginal distributions  each sum to one.

\begin{align} 
p_{000} + p_{100} &= p_\text{public}(0,0) \\ 
p_{010} + p_{110} &= p_\text{public}(1,0)  \\ 
p_{001} + p_{101} &= p_\text{public}(0,1) \\
(p_{000} +p_{001}+p_{100}+p_{101} )p_\text{bank}(0\mid0)&=p_{000}+p_{001},\\
(p_{010} +p_{011}+p_{110}+p_{111})p_\text{bank}(0\mid1)&=p_{010}+p_{011}\\
\sum_{s, o, e} p_{soe} = 1
\end{align}

\subsubsection*{Free Parameters}

This is an underdetermined linear system: 6 equations for 8 unknowns, hence there are 2 degrees of freedom. We can fix two parameters, say:

\begin{align}
    p_{000} = c, \quad p_{010} = k
\end{align}

\subsubsection*{Solving the Remaining Variables}

Given \(c\) and \(k\), we can compute all other variables:

\begin{align}
p_{000} &= c \\
p_{010} &= k \\
p_{100} &= p_\text{public}(0,0) - c \\
p_{110} &= p_\text{public}(1,0) - k \\
p_{001} &= p_\text{bank}(0 \mid 0) \cdot \left[p_\text{public}(0,0) + p_\text{public}(0,1)\right] - c \\
p_{101} &= p_\text{public}(0,1) - p_{001} \\
p_{011} &= p_\text{bank}(0 \mid 1) \cdot \left[p_\text{public}(1,0) + p_\text{public}(1,1)\right] - k \\
p_{111} &= p_\text{public}(1,1) - p_{011}
\end{align}

\subsubsection*{Parameter Bounds}

To ensure that all \(p_{soe} \geq 0\), we must impose bounds on \(c\) and \(k\):

For $c$

\begin{align}
\text{lower bound} &= \text{max}(0, p_\text{public}(0, 0)-1, p_\text{bank}(0 \mid 0) \cdot (p_\text{public}(0,0) + p_\text{public}(0,1)) - 1) \\
\text{upper bound} &= \text{min}(1, p_\text{public}(0, 0), p_\text{bank}(0 \mid 0) \cdot (p_\text{public}(0,0) + p_\text{public}(0,1)) 
\end{align}

Similarly for $k$

\begin{align}
\text{lower bound} &= \text{max}(0, p_\text{public}(1, 0)-1, p_\text{bank}(0 \mid 1) \cdot (p_\text{public}(1,0) + p_\text{public}(1,1)) - 1) \\
\text{upper bound} &= \text{min}(1, p_\text{public}(1, 0), p_\text{bank}(0 \mid 1) \cdot (p_\text{public}(1,0) + p_\text{public}(1,1)) 
\end{align}

\subsubsection*{Iterating Over Free Parameters}

We obtain the feasible set of joint distributions by iterating over a discretised grid of the two free parameters, \(c\) and \(k\) at 100 linearly spaced values within their valid bounds. This results in a total of \(100 \times 100 = 10{,}000\) joint distributions.

\subsection{Case For Consistent Marginals}

This is the case for when $p_\text{bank}(o) = p_\text{public}(o)$. For this case, the only difference lies in the constraints related to  $p_\text{bank}$, while the other constraints remain unchanged.

\begin{itemize}
    \item \textbf{Marginal constraints from \(p_\text{bank}(s, o)\):}
    \begin{align}
        p_{000} + p_{001} &= p_\text{bank}(0, 0) \\
        p_{100} + p_{101} &= p_\text{bank}(1, 0) \\
        p_{010} + p_{011} &= p_\text{bank}(0, 1) \\
        p_{110} + p_{111} &= p_\text{bank}(1, 1)
    \end{align}
\end{itemize}

However, we notice that the constraints in Section \ref{app:incost_marg} are equivalent to the above when $p_\text{bank}(o) = p_\text{public}(o)$.  Therefore we can use the same method as above.
In this situation, the joint distribution is automatically normalised because both marginals sum to one and are consistent. Thus, normalisation is redundant, leaving only the three non-redundant marginal constraints from each marginal data.

\section{General Case for Obtaining Feasible Set of Joint Distributions}
\label{app:general-case}

In the general case, our goal is to enable fairness evaluation by estimating the full joint distribution:
\begin{equation}
p(\textit{internal}, \textit{common}, \textit{external}),
\end{equation}
using only two incomplete marginal datasets with overlap between them $p_{\text{i}}(\textit{internal}, \textit{common})$ and $ p_{\text{e}}(\textit{common}, \textit{external})$. 

We let \textit{internal}, \textit{common}, and \textit{external} be sets of $n_i. n_o$ and $n_e$ binary variables respectively. Then the joint distribution \(p(\textit{internal}, \textit{common}, \textit{external})\) is defined over \(n = n_i + n_o + n_e\) binary variables and therefore has \(2^n\) parameters.

The joint distribution \( p(\textit{internal}, \textit{common}, \textit{external}) \) must satisfy the properties of a valid probability distribution. In particular, the full joint must be non-negative and normalised:
\begin{equation}
\sum_{\textit{internal}, \textit{common}, \textit{external}} p(\textit{internal}, \textit{common}, \textit{external}) = 1.
\end{equation}
Moreover, it is assumed that the marginal distributions are valid distributions  i.e.,
\begin{equation}
\sum_{\textit{internal, common}} p_i(\textit{internal}, \textit{common}) = 1, \quad \sum_{\textit{common, external}} p_e(\textit{common}, \textit{external}) = 1.
\end{equation}

\subsection{Case of Consistent Marginals}

 We require a joint distribution \(p(\textit{internal}, \textit{common}, \textit{external})\) satisfying:
\begin{align}
p_{\text{e}}(\textit{common}, \textit{external}) &= \sum_{\textit{internal}} p(\textit{internal}, \textit{common}, \textit{external}), \\ p_{\text{i}}(\textit{internal}, \textit{common}) &= \sum_{\textit{external}} p(\textit{internal}, \textit{common}, \textit{external}).
\end{align}

These marginals impose constraints on the joint distribution by requiring that their respective entries match the appropriate sums over the joint. Each marginal distribution over \(n_i + n_o\) or \(n_o + n_e\) binary variables has \(2^{n_i + n_o}\) and \(2^{n_o + n_e}\) entries, respectively. However, because each marginal must sum to 1, one of these constraints is redundant in each case. Therefore, the number of independent linear constraints contributed by the internal and external marginals are

\begin{equation}
2^{n_i + n_o} - 1 \quad \text{and} \quad 2^{n_o + n_e} - 1.
\end{equation}
Therefore, the total number of constraints is:
\begin{equation}
2^{n_i + n_o} + 2^{n_o + n_e} - 2.
\end{equation}
The number of free parameters in the joint distribution is:
\begin{align}
\#\text{free parameters} &= 2^{n} - (2^{n_i + n_o} + 2^{n_o + n_e} - 2) \nonumber \\
&= 2^{n} - 2^{n_i + n_o} - 2^{n_o + n_e} + 2.
\end{align}

\subsection{Case of Inconsistent Marginals}

 We require a joint distribution \(p(\textit{internal}, \textit{common}, \textit{external})\) satisfying:
\begin{align}
p_{\text{e}}(\textit{common}, \textit{external}) &= \sum_{\textit{internal}} p(\textit{internal}, \textit{common}, \textit{external}), \\ p_{\text{i}}(\textit{internal} \mid \textit{common}) &= \frac{\sum_{\textit{external}} p(\textit{internal}, \textit{common}, \textit{external})}{\sum_{\textit{internal}, \textit{external}} p(\textit{internal}, \textit{common}, \textit{external})}..
\end{align}

For each of the \(2^{n_o}\) settings of \(\textit{common}\), the conditional distribution \(p(\textit{internal} \mid \textit{common})\) specifies a distribution over \(2^{n_i}\) outcomes, subject to a normalisation constraint. Thus, for each setting we get \(2^{n_i} - 1\) degrees of freedom.  When marginals are consistent and fully specified, normalisation is automatically ensured, so the normalisation constraint itself becomes redundant. However, if marginals are inconsistent or incomplete, normalisation can’t be inferred and must be included explicitly, making fewer constraints redundant. Therefore, the number of independent linear constraints contributed by the internal and external marginals and normalisation are

\begin{equation}
2^{n_o}(2^{n_i} - 1)  \quad \text{and} \quad 2^{n_o + n_e} - 1  \quad \text{and} \quad 1
\end{equation}

Therefore, the total number of constraints is:

\begin{align}
\text{Total constraints} &= 2^{n_o}(2^{n_i} - 1) + 2^{n_o + n_e} - 1 + 1\nonumber \\
&= 2^{n_o + n_e} + 2^{n_o}(2^{n_i} - 1) .
\end{align}

The number of free parameters in the joint distribution is:

\begin{align}
\#\text{free parameters} &= 2^n - \left(2^{n_o + n_e} + 2^{n_o}(2^{n_i} - 1) \right) \nonumber \\
&= 2^n - 2^{n_o + n_e} - 2^{n_i + n_o} + 2^{n_o}.
\end{align}

\subsection{Computational Cost}

In our experiments, we iterated over 100 values for each free parameter. For the case of 3 binary variables, there are 2 free parameters, resulting in \(100^2 = 10^{4}\) combinations to explore in order to obtain the feasible set. Running the iterative algorithm over this grid to generate the list of feasible joint distributions takes approximately 80 milliseconds on the CPU integrated with M3 Max processor. However, this approach quickly becomes intractable as the number of variables increases. For instance, when \(n_i = 2\), \(n_o = 1\), and \(n_e = 2\), the joint distribution contains \(2^5 = 32\) entries, and the constraints leave 18 free parameters. Iterating over a grid with 100 values per parameter would require \(100^{18} = 10^{36}\) combinations.

\section{Details for Expectation Maximisation Algorithm for Latent Naïve Bayes Method}
\label{app:em-algo}

We aim to estimate the joint distribution from two marginal datasets (internal and external data) which some overlap between them. To do this, we use a structural assumption called the Latent Na\"ive Bayes model, which assumes that all observed variables are conditionally independent given a latent variable $Z$.

\subsection{Setup}  
Let there be categorical variables 
\begin{equation}
X_1, X_2, \ldots, X_{I+J+L}
\end{equation}
with respective domains 
\begin{equation}
\mathrm{dom}(X_i) = \{1, 2, \ldots, M_i\},
\end{equation}
where \(M_i \geq 2\). The variables are partitioned into:
\begin{itemize}
    \item Internal variables observed only in dataset \(\mathbf{D_1}\): indices \(i \in \{1, \ldots, I\}\),
    \item External variables observed only in dataset \(\mathbf{D_2}\): indices \(i \in \{I+1, \ldots, I+J\}\),
    \item Common overlapping variables observed in both datasets: indices \(i \in \{I+J+1, \ldots, I+J+L\}\).
\end{itemize}

\paragraph{Datasets:}  
\begin{equation}
\mathbf{D_1} = \{ (x_1^n, \ldots, x_I^n, x_{I+J+1}^n, \ldots, x_{I+J+L}^n) \}_{n=1}^{N_1}
\end{equation}
\begin{equation}
\mathbf{D_2} = \{ (x_{I+J+1}'^n, \ldots, x_{I+J+L}'^n, x_{I+1}^n, \ldots, x_{I+J}^n) \}_{n=1}^{N_2}
\end{equation}

\paragraph{Latent variable:}  
Introduce a latent variable \(Z\) with
\begin{equation}
Z \in \{1, 2, \ldots, K\}.
\end{equation}

\paragraph{Parameters:}

\begin{itemize}
    \item \textbf{Latent variable probabilities:}
    \begin{equation}
    \pi_k = p(Z = k), \quad k = 1, 2, \ldots, K,
    \end{equation}
    which satisfy
    \begin{equation}
    \pi_k \geq 0, \quad \text{and} \quad \sum_{k=1}^K \pi_k = 1.
    \end{equation}

    \item \textbf{Conditional distributions of observed variables:}  
    For each observed variable \(X_i\), and each category \(k\), the model specifies
    \begin{equation}
    p_i(m \mid k) = p(X_i = m \mid Z = k), \quad m = 1, 2, \ldots, M_i,
    \end{equation}
    where \(M_i\) is the number of possible values \(X_i\) can take. These probabilities satisfy
    \begin{equation}
    p_i(m \mid k) \geq 0, \quad \text{and} \quad \sum_{m=1}^{M_i} p_i(m \mid k) = 1.
    \end{equation}
\end{itemize}

Learning the model corresponds to estimating the parameters \(\{\pi_k\}\) and \(\{p_i(m \mid k)\}\) from the data. Once these parameters are estimated, the joint distribution can be computed as
\begin{equation}
p_{\boldsymbol{\theta}}(X_1, \ldots, X_n) = \sum_{k=1}^K \pi_k \prod_{i=1}^n p_i(X_i \mid k).
\end{equation}

\paragraph{Observed data log-likelihoods:}

Define index sets
\begin{equation}
\mathcal{I}_1 = \{1,\ldots,I\} \cup \{I+J+1, \ldots, I+J+L\},
\end{equation}
\begin{equation}
\mathcal{I}_2 = \{I+1, \ldots, I+J\} \cup \{I+J+1, \ldots, I+J+L\},
\end{equation}
\begin{equation}
\mathcal{I}_{\text{common}} = \{I+J+1, \ldots, I+J+L\},
\quad
\mathcal{I}_2^{\text{ext}} = \mathcal{I}_2 \setminus \mathcal{I}_{\text{common}} = \{I+1, \ldots, I+J\}.
\end{equation}

\begin{equation}
\log p_{\boldsymbol{\theta}}(\mathbf{D_1}) = \sum_{n=1}^{N_1} \log \left( \sum_{k=1}^K \pi_k \prod_{i \in \mathcal{I}_1} p_i(x_i^n \mid k) \right),
\end{equation}
\begin{equation}
\log p_{\boldsymbol{\theta}}(\mathbf{D_2}) = \sum_{n=1}^{N_2} \log \left( \sum_{k=1}^K \pi_k \left( \prod_{i \in \mathcal{I}_2^{\text{ext}}} p_i(x_i'^n \mid k) \right) \left( \prod_{i \in \mathcal{I}_{\text{common}}} p_i(x_i^n \mid k) \right) \right).
\end{equation}

\paragraph{Evidence Lower Bound (ELBO):}  
For the two datasets, introduce variational distributions \( q_1(Z) \) and \( q_2(Z) \) over the latent variable \(Z\). Using Jensen's inequality, the observed data log-likelihoods satisfy:

\begin{equation}
\log p_{\boldsymbol{\theta}}(\mathbf{D}_1) 
\geq \sum_{n=1}^{N_1} \sum_{k=1}^K q_1(k) \left[ \log \pi_k + \sum_{i \in \mathcal{I}_1} \log p_i(x_i^n \mid k) - \log q_1(k) \right].
\end{equation}

Similarly, for \(\mathbf{D}_2\):
\begin{equation}
\log p_{\boldsymbol{\theta}}(\mathbf{D}_2) \geq \sum_{n=1}^{N_2} \sum_{k=1}^K q_2(k) \left[
\log \pi_k + \sum_{i \in \mathcal{I}_2^{\text{ext}}} \log p_i(x_i'^n \mid k) + \sum_{i \in \mathcal{I}_{\text{common}}} \log p_i(x_i^n \mid k) - \log q_2(k)
\right].
\end{equation}

Combining both, the total ELBO is:

\begin{align}
\mathcal{L}(\boldsymbol{\theta}, q_1, q_2) &= \sum_{n=1}^{N_1} \sum_{k=1}^K q_1(k) \left[ \log \pi_k + \sum_{i \in \mathcal{I}_1} \log p_i(x_i^n \mid k) - \log q_1(k) \right] \\
&\quad + \sum_{n=1}^{N_2} \sum_{k=1}^K q_2(k) \left[ \log \pi_k + \sum_{i \in \mathcal{I}_2^{\text{ext}}} \log p_i(x_i'^n \mid k) + \sum_{i \in \mathcal{I}_{\text{common}}} \log p_i(x_i^n \mid k) - \log q_2(k) \right].
\end{align}

\subsection{EM Algorithm Steps}

The EM algorithm alternates between two steps to maximise the ELBO:

\begin{itemize}
    \item \textbf{E-step:} With the current model parameters \(\boldsymbol{\theta}^{(t-1)}\) fixed, update the variational distributions \(q_1^{(t)}\) and \(q_2^{(t)}\) over the latent variables to maximise the ELBO.
    \item \textbf{M-step:} With the updated variational distributions \(q_1^{(t)}\) and \(q_2^{(t)}\) fixed, maximise the ELBO with respect to the model parameters \(\boldsymbol{\theta}^{(t)}\).
\end{itemize}

By iterating these steps, the ELBO improves monotonically until convergence.

\paragraph{E-step: Responsibilities}  
For each \(n\) and \(k\),
\begin{equation}
q_1^{(t)}(z^n = k) = \frac{\pi_k^{(t-1)} \prod_{i \in \mathcal{I}_1} p_i^{(t-1)}(x_i^n | k)}{\sum_{j=1}^K \pi_j^{(t-1)} \prod_{i \in \mathcal{I}_1} p_i^{(t-1)}(x_i^n | j)},
\end{equation}
\begin{equation}
q_2^{(t)}(z'^n = k) = \frac{
\pi_k^{(t-1)} 
\left(\prod_{i \in \mathcal{I}_2^{\text{ext}}} p_i^{(t-1)}(x_i^n \mid k)\right)
\left(\prod_{i \in \mathcal{I}_{\text{common}}} p_i^{(t-1)}(x_i'^n \mid k)\right)
}{
\sum_{j=1}^K \pi_j^{(t-1)} 
\left(\prod_{i \in \mathcal{I}_2^{\text{ext}}} p_i^{(t-1)}(x_i^n \mid j)\right)
\left(\prod_{i \in \mathcal{I}_{\text{common}}} p_i^{(t-1)}(x_i'^n \mid j)\right)
}.
\end{equation}

---

\paragraph{M-step: Parameter updates}

\paragraph{Mixing proportions:}

\begin{equation}
\pi_k^{(t)} = \frac{\sum_{n=1}^{N_1} q_1^{(t)}(z^n = k) + \sum_{n=1}^{N_2} q_2^{(t)}(z'^n = k)}{N_1 + N_2}.
\end{equation}

\paragraph{Conditional probabilities:}

For each variable \(i\) with domain \(\mathcal{M}_i\), the updates are:

\begin{equation}
p_i^{(t)}(m|k) = \frac{S_i(m,k)}{\sum_{m' \in \mathcal{M}_i} S_i(m',k)},
\end{equation}

where

\begin{equation}
S_i(m, k) = 
\begin{cases}
\sum_{n=1}^{N_1} \mathbf{1}[x_i^n = m] \, q_1^{(t)}(z^n = k), & \text{if } i \in \mathcal{I}_1 \\
\sum_{n=1}^{N_2} \mathbf{1}[x_i^n = m] \, q_2^{(t)}(z'^n = k), & \text{if } i \in \mathcal{I}_2^{\text{ext}} \\
\sum_{n=1}^{N_1} \mathbf{1}[x_i^n = m] \, q_1^{(t)}(z^n = k) + \sum_{n=1}^{N_2} \mathbf{1}[x_i'^n = m] \, q_2^{(t)}(z'^n = k), & \text{if } i \in \mathcal{I}_{\text{common}}
\end{cases}
\end{equation}

\section{Dataset Details}
\label{app:real_data_details}

For the Adult Data, the `fnlwgt' attribute is dropped as it is not relevant to the task and the `education-num' attribute as it duplicates the information available in the `education' attribute. COMPAS Data is filtered to only include `race` column is either `African-American' or `Caucasian' and coding as $\{black, white\}$. We further combine three columns containing juvenile crime counts to get the total number of juvenile crimes. Details of the attributes and their values can be found in Tables \ref{tab:adult}, \ref{tab:compas}, and \ref{tab:german_credit}.

\begin{table}[h]
  \caption{Adult Data: Attributes and Their Values}
  \label{tab:adult}
  \centering
  \begin{tabular}{lp{10cm}}
    \toprule
    \textbf{Attribute} & \textbf{Values} \\
    \midrule
    Age & \{25--60, \textless 25, \textgreater 60\} \\
    Capital Gain & \{\textless=5000, \textgreater 5000\} \\
    Capital Loss & \{\textless=40, \textgreater 40\} \\
    Education & \{assoc-acdm, assoc-voc, bachelors, doctorate, HS-grad, masters, prof-school, some-college, high-school, primary/middle school\} \\
    Hours Per Week & \{\textless 40, 40--60, \textgreater 60\} \\
    Income & \{\textless=50K, \textgreater 50K\} \\
    Marital Status & \{married, other\} \\
    Native Country & \{US, non-US\} \\
    Occupation & \{adm-clerical, armed-forces, craft-repair, exec-managerial, farming-fishing, handlers-cleaners, machine-op-inspct, other-service, priv-house-serv, prof-specialty, protective-serv, sales, tech-support, transport-moving\} \\
    Race & \{non-white, white\} \\
    Relationship & \{non-spouse, spouse\} \\
    Sex & \{male, female\} \\
    Workclass & \{private, non-private\} \\
    \bottomrule
  \end{tabular}
\end{table}

\begin{table}[h]
  \caption{COMPAS Data: Attributes and Their Values}
  \label{tab:compas}
  \centering
  \begin{tabular}{lp{10cm}}
    \toprule
    \textbf{Attribute} & \textbf{Values} \\
    \midrule
    Age Category & \{25 - 45, \textgreater 45, \textless 25\} \\
    Charge Degree & \{F, M\} \\
    Juvenile Crime & \{0, 1, 2, 3, 4, 5, 6, 7, 8, 9, 10, 11, 14\} \\
    Priors Count & \{0, 1, 2, 3, 4, 5, 6, 7, 8, 9, 10, 11, 12, 13, 14, 15, 16, 17, 18, 19, 20, 21, 22, 23, 24, 25, 26, 27, 28, 29, 30, 31, 33, 36, 37, 38\} \\
    Race & \{Black, White\} \\
    Score Text & \{High, Low, Medium\} \\
    Sex & \{Female, Male\} \\
    Two-Year Recidivism & \{0, 1\} \\
    Violent Score Text & \{High, Low, Medium\} \\
    \bottomrule
  \end{tabular}
\end{table}

\begin{table}[h]
  \caption{German Credit Data: Attributes and Their Values}
  \label{tab:german_credit}
  \centering
  \begin{tabular}{p{3cm}p{10cm}}
    \toprule
    \textbf{Attribute} & \textbf{Values} \\
    \midrule
    Age & \{<= 25, >25\} \\
    Checking Account & \{0 <= <200 DM, <0 DM, >= 200 DM, no account\} \\
    Class Label & \{0, 1\} \\
    Credit Amount & \{<=2000, 2001-5000, >5000\} \\
    Credit History & \{all credits at this bank paid back duly, critical account, delay in paying off, existing credits paid back duly till now, no credits taken\} \\
    Duration & \{<=6, 7-12, >12\} \\
    Employment Since & \{1 <= < 4 years, 4 <= <7 years, <1 years, >=7 years, unemployed\} \\
    Existing Credits & \{1, 2, 3, 4\} \\
    Foreign Worker & \{no, yes\} \\
    Housing & \{for free, own, rent\} \\
    Installment Rate & \{1, 2, 3, 4\} \\
    Job & \{management/ highly qualified employee, skilled employee / official, unemployed/ unskilled - non-resident, unskilled - resident\} \\
    Marital Status & \{divorced/separated, married/widowed\} \\
    Number of People Provide Maintenance For & \{1, 2\} \\
    Other Debtors & \{co-applicant, guarantor, none\} \\
    Other Installment Plans & \{bank, none, store\} \\
    Property & \{car or other, real estate, savings agreement/life insurance, unknown / no property\} \\
    Purpose & \{business, car (new), car (used), domestic appliances,
education, furniture/equipment, others, radio/television,
repairs, retraining\} \\
    Residence Since & \{1, 2, 3, 4\} \\
    Savings Account & \{100 <= <500 DM, 500 <= < 1000 DM, <100 DM, >= 1000 DM, no savings account\} \\
    Sex & \{female, male\} \\
    Telephone & \{none, yes\} \\
    \bottomrule
  \end{tabular}
\end{table}

\section{Real Data Experiment Details}
\label{app:real_data_exp}

\subsection{Evaluating Feasible Set Method}
To apply the feasible set method to real-world datasets, we reduced each to three binary variables and partitioned them into internal and external datasets, as described below. These provide the marginal datasets which we use to obtain feasible set of joint distributions and possible fairness values.

In the German Credit dataset, the internal dataset included \texttt{Employment Since} (at least 4 years / less than 4 years) and \texttt{Housing} (owns / does not own), while the external dataset included \texttt{Housing} and protected attribute \texttt{Sex} (male / female).

For the Adult Income dataset, the internal dataset included  \texttt{Capital Gain} (high / low) and \texttt{Marital Status} (married / not married), while the external dataset included \texttt{Marital Status} and protected attribute \texttt{Sex} (male / female).

In the COMPAS dataset, the internal dataset included  \texttt{Priors Count} (less than 5 / 5 or more) and \texttt{Risk Score} (low / medium or high), while the external dataset includes  \texttt{Risk Score}  and protected attribute \texttt{Race} (African-American / Caucasian).

In all cases, the class label was retained and used to train a Decision Tree Classifier to predict based on variables in the internal data.

\subsubsection{Additional Results}

\begin{figure}[h]
\centering
\begin{subfigure}[b]{0.9\textwidth}
    \includegraphics[width=\textwidth]{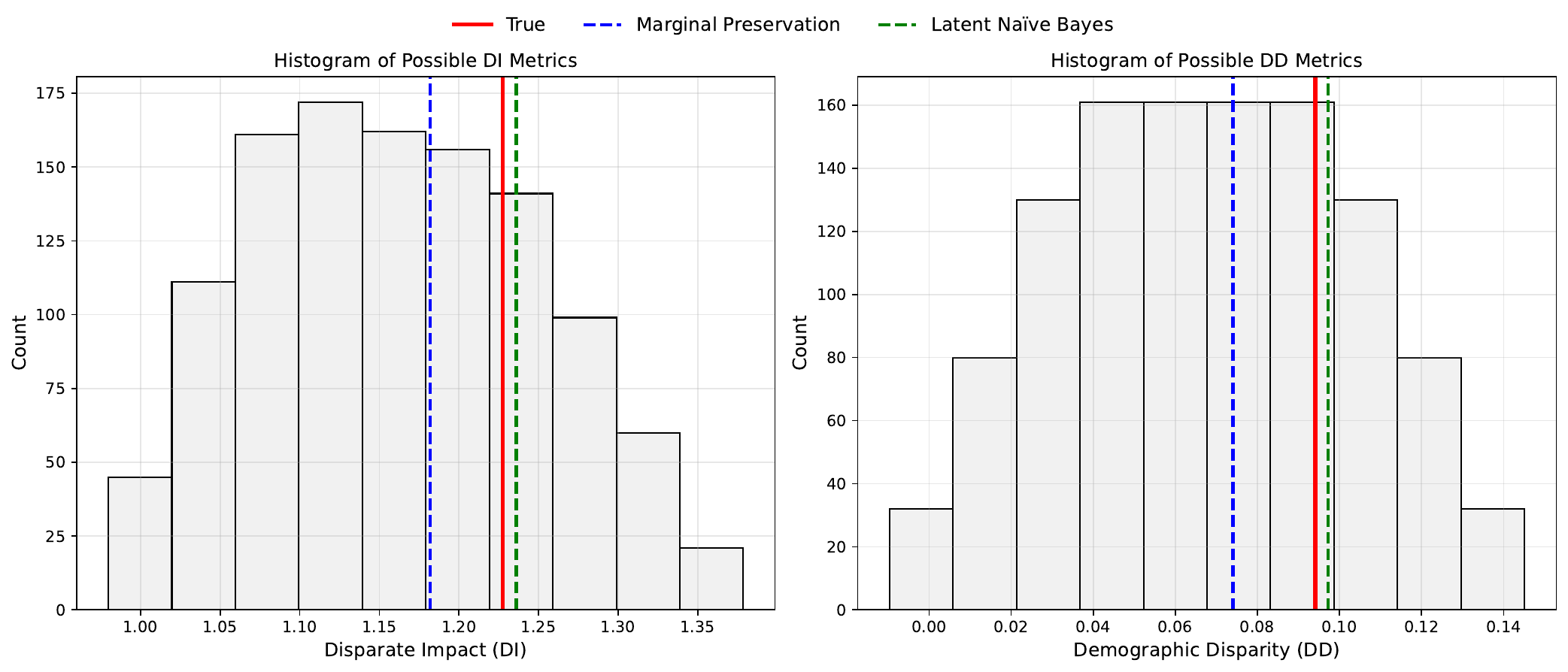}
    \caption{Results using COMPAS data.}
    \label{fig:real_hists_compas}
\end{subfigure}

\vspace{1em} 

\begin{subfigure}[b]{0.9\textwidth}
    \includegraphics[width=\textwidth]{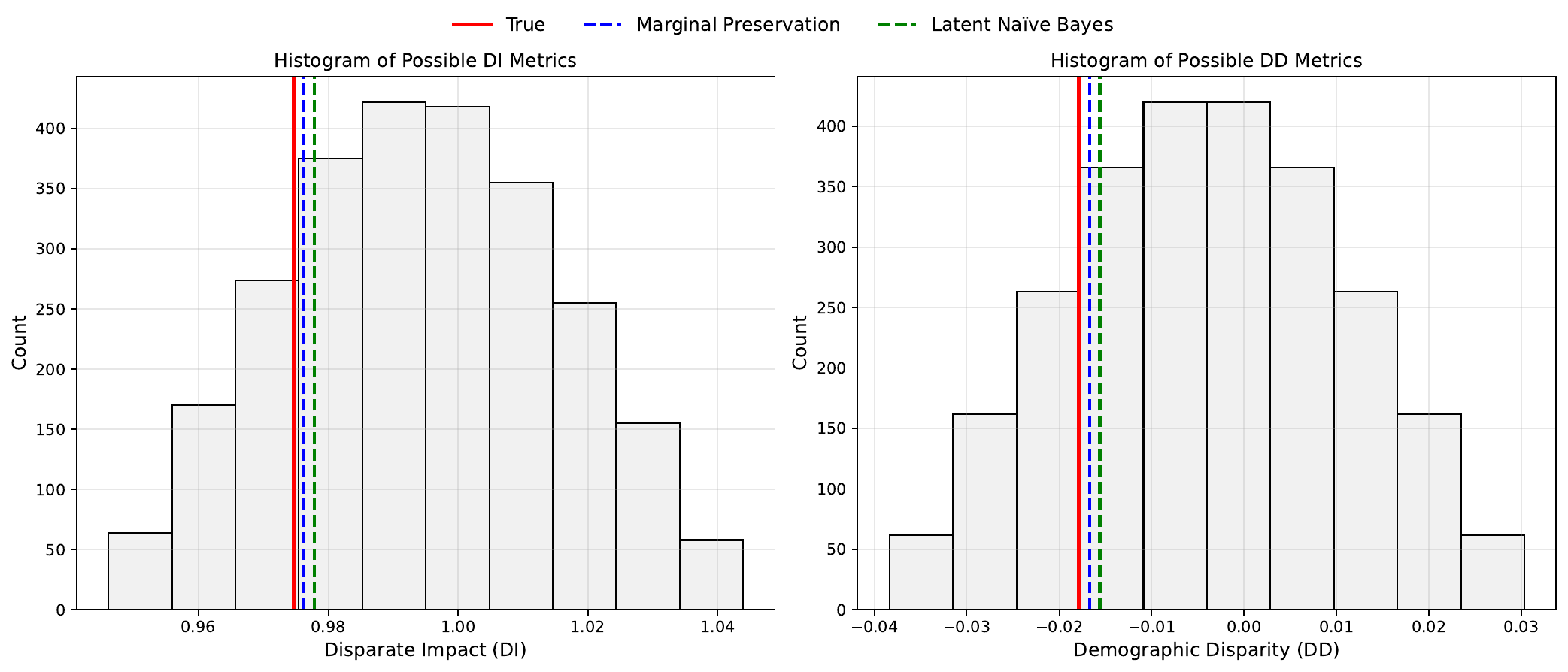}
    \caption{Results using German Credit data.}
    \label{fig:real_hists_german}
\end{subfigure}

\caption{Distribution of possible values for a fairness metric, Disparate Impact (DI) or Demographic Disparity (DD), computed from the feasible set of joint distributions consistent with observed marginals. The true fairness metric (from the full joint) is marked in red. Fairness estimates from two structural assumption-based methods are shown in green and blue for comparison.}
\label{fig:real_hists_combined}
\end{figure}

Figures~\ref{fig:real_hists_compas} and~\ref{fig:real_hists_german} provide additional results not shown in the main text. These plots illustrate the distribution of fairness metrics, Disparate Impact (DI) or Demographic Disparity (DD), computed from the feasible set of joint distributions consistent with internal and external marginals data for the COMPAS and German Credit datasets, respectively. They also compare fairness estimates obtained using structural assumption-based methods.

\subsection{Evaluating Structural Assumption Method}

We retained all variables and  multivariate categories in the processed real data. Simulated partitions of internal and external data are shown in Table~\ref{tab:separate-data}. 

In all cases, the class label was preserved and used to train a decision tree classifier based solely on variables from the internal dataset. The data used to train the classifier was distinct from the marginal data used to estimate the joint distribution. Additionally, a hold-out test set (30\% of full data) was used as ground truth joint distribution. This allowed us to compare fairness metrics computed using the classifier and the true joint distribution against those computed using the classifier and the estimated joint distribution using the structural assumption methods.

Rather than representing joint distributions explicitly as large probability tables, which becomes infeasible for datasets with many categorical variables and a large number of possible combinations, we instead sample directly from the data. This sampling-based approach is equivalent to using the empirical distribution, where probabilities are implicitly defined by observed frequencies. This avoids the need to store or manipulate extremely small probability values, which can be computationally inefficient and numerically unstable in large-scale settings.

In our framework, once the joint distribution is estimated, either through Latent Naïve Bayes or Marginal Preservation, we generate bootstrap samples accordingly. Each bootstrap sample consists of the same number of instances as the real test dataset and is drawn according to the probabilities in the estimated joint. These samples are then used to compute fairness metrics, such as Disparate Impact or Demographic Disparity, by evaluating the classifier's predictions on the sampled data.

By repeating this process across 1,000 bootstrap iterations, we can obtain a mean over fairness metrics that accounts for variability introduced by sampling from the estimated joint distribution. We compare the average (mean) fairness metric across the 1,000 bootstrap samples, with the fairness metric computed using the true joint distribution from the actual data. The results show the absolute difference between these two as your result, this quantifies how close the estimated fairness is to the ground truth.

\begin{table}[h]
\caption{Separation of complete real datasets, with each row illustrating how attributes are categorised into external and internal datasets. The external dataset shown includes demographic protected attributes, while the internal dataset comprises the remaining attributes. Overlapping variables shared between the two datasets are shown in \textit{italics}.}
\label{tab:separate-data}
\begin{center}
\begin{tabular}{ll}
\toprule
    \textbf{Dataset} & \textbf{Attributes in External Dataset (overlapping variable in \textit{italics})}  \\
    \midrule
    Adult  
    & \textit{marital-status}, age, sex, race, relationship, native-country \\
    COMPAS   
    & \textit{score}, sex, age, race \\
    German Credit  
    & \textit{housing}, sex, marital-status, age, foreign-worker \\
\bottomrule
\end{tabular}
\end{center}
\vspace{-1em}
\end{table}

\subsubsection{Additional Results}

Additional results for the German Credit dataset, not included in the main text, are presented in Table~\ref{tab:german_di_only}.

\begin{table}[h]
  \caption{Absolute difference in Disparate Impact (DI) between fairness metrics derived from using joint estimated using two structural independence assumptions (Latent Bayes, and Marginal Preservation) vs fairness metric from the true joint for a Decision Tree Classifier.}
  \label{tab:german_di_only}
  \centering
  \begin{tabular}{llc}
    \toprule
    \textbf{Dataset – Attribute} & \textbf{Method } & \textbf{DI diff} $\downarrow$ \\
    \midrule
    German – Sex & Latent  & 0.190 \\
                & Marginal & 0.178 \\

    \bottomrule
  \end{tabular}
\end{table}